\pdfoutput=1

\documentclass[11pt]{article}

\usepackage[preprint]{acl}

\usepackage{times}
\usepackage{latexsym}

\usepackage[T2A,T1]{fontenc}

\usepackage[utf8]{inputenc}

\usepackage{microtype}

\usepackage{inconsolata}

\usepackage{booktabs}
\usepackage{graphicx}
\usepackage{float}
\usepackage{caption}
\usepackage{enumitem}

\title{{T}artu{NLP} @ {AXOLOTL}-24: {L}everaging {C}lassifier {O}utput for {N}ew {S}ense {D}etection in {L}exical {S}emantics}

\author{Aleksei Dorkin \and Kairit Sirts \\
        Institute of Computer Science \\
        University of Tartu \\
        \texttt{\{aleksei.dorkin, kairit.sirts\}@ut.ee} \\ }

\begin{document}
\maketitle
\begin{abstract}

We present our submission to the AXOLOTL-24 shared task. The shared task comprises two subtasks: identifying new senses that words gain with time (when comparing newer and older time periods) and producing the definitions for the identified new senses. We implemented a conceptually simple and computationally inexpensive solution to both subtasks. We trained adapter-based binary classification models to match glosses with usage examples and leveraged the probability output of the models to identify novel senses. The same models were used to match examples of novel sense usages with Wiktionary definitions. Our submission attained third place on the first subtask and the first place on the second subtask.

\end{abstract}

\section{Introduction}

The subject of the AXOLOTL-24 shared task~\cite{fedorova-etal-2024-axolotl} is diachronic semantic change detection and explanation. Diachronic semantic change is understood as the change in word meanings (i.e., words losing old senses and obtaining new ones) over shorter or longer periods. Accordingly, given a dataset containing usage examples from different periods (old and new), the task is to identify and define the new senses that words gain in the new time period compared to the old one.

The goal of the shared task is to implement a semantic change modeling system for two tasks:

\begin{enumerate}[label=\arabic*),noitemsep]
    \item Correctly assigning existing senses to target word usages and identifying novel, previously unseen senses;
    \item Describing the identified novel senses.
\end{enumerate}

The data in the shared task is provided in three languages: Finnish, Russian, and German (the surprise language for which only the test split is available). For each language, examples from old and new periods are given. Each data point consists of a target word and its usage example, gloss (target word definition), the period the example comes from, and usage and sense IDs. The data includes glosses for both time periods in the training and validation splits, while glosses for the new time period are not provided in the test splits.
The ``old'' and ``new'' periods differ for each language. For Finnish, old texts are dated before 1700, and new ones are dated after 1700. For Russian, the old target word usages are from the 19th century, and the new data represents modern usages of words. For German, the old period is from 1800 to 1899, and the new period is from 1946 to 1990~\cite{schlechtweg2023human}.

\begin{table}[t]
\small
\centering
\begin{tabular}{lcc}
\toprule
\textbf{Team} & \textbf{ARI} & \textbf{F1} \\
\midrule
deep-change    & \textbf{0.413} & \textbf{0.750}  \\
Holotniekat     & 0.312 & 0.641  \\
\textit{TartuNLP (ours)}        & 0.310 & 0.590  \\ 
IMS\_Stuttgart   & 0.287 & 0.487  \\
ABDN-NLP        & 0.221 & 0.431 \\
WooperNLP       & 0.187 & 0.316  \\
Baseline        & 0.041 & 0.207  \\
\bottomrule
\end{tabular}
\caption{Overall results on the Subtask 1.}
\label{tab:task1-all}
\end{table}

\begin{table}[t]
\small
\centering
\begin{tabular}{lccc}
\toprule
\textbf{Team} & \textbf{Overall} & \textbf{BLEU} & \textbf{BERTScore} \\ 
\midrule
\textit{TartuNLP (ours)}    & \textbf{0.467} & \textbf{0.208} & \textbf{0.726} \\
WooperNLP                   & 0.340          & 0.020          & 0.660          \\ 
ABDN-NLP                    & 0.253          & 0.045          & 0.461          \\ 
baseline                    & 0.218          & 0.013          & 0.423          \\
\bottomrule
\end{tabular}
\caption{Overall results on the Subtask 2.}
\label{tab:task2-all}
\end{table}

Although we participated in both subtasks, we were primarily interested in the second subtask of producing definitions for new senses. We implemented a solution that matches identified novel sense usages with definitions from an external resource (Wiktionary). Our approach is based on a binary classification task to predict whether a proposed definition matches the sense under consideration. We reused this binary classification model for the second subtask of describing the identified novel senses.

Our system attained the first place on the second subtask (Table \ref{tab:task2-all}) and obtained competitive results on the first subtask (Table \ref{tab:task1-all}).

\section{Methodology}

We propose a simple classification-based solution for both subtasks. We adopt the GlossBERT approach \cite{huang-etal-2019-glossbert} that treats word sense disambiguation as a sentence pair classification task, where each pair comprises a usage example and a sense definition. In turn, we frame the problem of new sense identification as the problem of matching between usage examples and sense definitions. Accordingly, the matching problem can be solved with a binary classification model that, given a usage example and a sense definition, outputs the probability of the sense definition correctly describing the usage example.

We adopt the cross-encoder model that simultaneously processes the usage examples and the sense definitions with the same model.
Given a usage example and a sense inventory, we apply the classification model to predict binary probabilities for each example/sense definition combination.
If the highest probability over all candidate pairs exceeds a predefined threshold, the system assigns the highest probability sense to the usage example.
Otherwise, the sense used in the example is deemed to be new.

\subsection{Subtask 1: Bridging Diachronic Word Uses and a Synchronic Dictionary}

This subtask aims to assign a sense ID to every usage example from the new period; the sense ID may come either from the senses in the old period or, if the system identifies a novel sense, a unique new sense ID is created.

The data for the first subtask contains sense definitions and correct usages, which can be used to construct positive examples for our task formulation. However, having only positive examples for a classification model is generally insufficient. To produce negative examples, we employ a simple algorithm. We only consider words associated with at least two distinct sense IDs. For a given sense ID and its associated gloss, we create all possible combinations of the gloss with the usage examples associated with the other sense IDs of the same word (consult Appendix~\ref{appendix} for additional details). We expect that the negatives obtained with this algorithm are hard and, as such, are more useful for training that could be obtained, for instance, via random sampling.

We transform every split of every language by extending it with negative examples created with the procedure described above. For each language, we train a separate classification model on the train split and evaluate on the development split. When training and evaluating the classifier, we do not consider the period (old or new) from which the examples come. The best checkpoint for each model is selected based on the development F1 score.

Having trained the classification models, we perform inference on the test set and transform the output into the expected format. During inference, 
the usage examples from the old period are ignored and the classification is performed only on pairs of usage examples from the new period and sense definitions from the old period. If the highest predicted probability for a usage example is above a threshold, we assign the sense ID of the most probable sense definition to the usage example. Otherwise, a new sense ID is created. The final result submitted for evaluation contains both the predicted senses for the examples from the new period as well as the positive examples in the test split from the old period.\footnote{Since the test examples for the old period were already annotated, we simply copied their sense definitions to the submitted result file.}

For the surprise language---German---the training process is slightly different. No training or validation data is provided, so we train and validate the classification model on the positive and negative examples obtained from the old period in the test data. Inference, however, is exactly the same.

\subsection{Subtask 2: Definition Generation for Novel Word Senses}

Subtask 2 aims to define each novel sense identified in the first subtask. Despite the name of the subtask, our approach does not generate any new definitions. We also do not train any additional models. As previously mentioned, we consider this task a matching problem, except that the definitions for the novel senses are not present in the data provided in the shared task. To solve this problem, we scrape the definitions of the surface forms, for which we identified at least one example as the usage of a novel sense, from the Wiktionary. More specifically, we scrape the definitions from the language-specific Wiktionary versions for each language (i.e. Finnish,\footnote{\url{https://fi.wiktionary.org/}} Russian,\footnote{\url{https://ru.wiktionary.org/}} and German\footnote{\url{https://de.wiktionary.org/}} Wiktionaries).

Having scraped the necessary definitions, we head straight to inference on the test set. We reuse the models and predictions from the first subtask. We collect the examples identified as the usages of the novel senses from the predictions and match them with the Wiktionary definitions using the classifier models trained in the first subtask. After that, we add the matched definitions to the predicted new senses.

\subsection{Implementation Details}

Different from GlossBERT~\cite{huang-etal-2019-glossbert}, which is based on the BERT~\cite{devlin-etal-2019-bert} model, we instead use XLM-RoBERTa~\cite{conneau2019unsupervised} as the base model for our classifiers. 
XLM-RoBERTa is a multilingual model that includes Finnish, Russian, and German in its training data. 
We expect our system to benefit from the multilinguality. 
Instead of full fine-tuning, we opt for parameter-efficient fine-tuning.
More specifically, we train bottleneck adapter~\cite{houlsby2019parameter} classifiers for each language. We adopted this approach because it makes our solution computationally lightweight and easily reproducible.

In GlossBERT, \citet{huang-etal-2019-glossbert} differentiate between training setups with and without weak supervision, with the former including the defined word itself in the gloss, as well as highlighting it in the usage example. According to the experimental results reported by \citet{huang-etal-2019-glossbert}, weak supervision appears to bring minor improvements in sense prediction. However, we do not use weak supervision in our submission. The reason is that Finnish and Russian are substantially more morphologically rich languages than English; thus, the target words rarely appear in their dictionary forms in the usage examples. Moreover, in some cases, the orthography also differs between old and new periods.

To delimit context and gloss, \citet{huang-etal-2019-glossbert} use the special [SEP] token that is pre-trained into the BERT model via the next sentence prediction task. However, RoBERTa~\cite{liu2019roberta}, and by extension XLM-RoBERTa, omitted the next sentence prediction task in the pre-training. As a result of that, the </s> token that is used by RoBERTa in place of [SEP] does not have the same classification-oriented meaning. For this reason, we employed the tabulation symbol as the delimiter instead.

For each language, we employed different variations of the base model and varying training setups. For Finnish, we used the large version of XLM-RoBERTa and trained for ten epochs in half-precision with a batch size of 128 and 3 steps of gradient accumulation. We observed that the training did not converge with a smaller effective batch size.
For Russian, we trained the classifier adapter with the base version of XLM-RoBERTa for 50 epochs with a batch size of 144. We also experimented with the large version of the model for the Russian language; however, it showed no improvements compared to the base version.
For German, we did not train the classifier from scratch. Instead, we continued training from the best checkpoint trained on the Finnish data.
The motivation is that there is considerably more data in Finnish than in Russian in the shared task, so we assume the Finnish model to be stronger. 
We continued training the Finnish classifier for 20 epochs in half-precision with a batch size of 48 and 6 steps of gradient accumulation. All models were trained with a 5e-4 learning rate.

The threshold value for the classifier's probability to identify novel senses was selected as the highest scoring option in the first subtask using the evaluation script provided by the organizers. We tested a small number of values in the range of of 0.2 to 0.5 on Russian and determined the best value to be \textbf{0.35}. The same value was used for all languages without additional testing due to time limitations.

The models were trained on the University High-Performance Cluster~\cite{https://doi.org/10.23673/ph6n-0144}.
We used a single Tesla V100 GPU for Russian and German, while for Finnish, we used a single A100 80GB GPU. The time elapsed on training is 9 hours for Finnish, 3 hours for Russian, and 9 minutes for German.
We implemented our solution using the transformers\footnote{\url{https://github.com/huggingface/transformers}} and the adapters\footnote{\url{https://github.com/adapter-hub/adapters}} libraries. The source code and the data are available on GitHub\footnote{\url{https://github.com/slowwavesleep/ancient-lang-adapters/tree/axolotl}}
and HuggingFace Hub,\footnote{\url{https://huggingface.co/datasets/adorkin/axolotl-wiktionary-definitions}} respectively.

\begin{table*}[ht]
\small
\centering
\begin{tabular}{lcccccc}
\hline
Team & Fi-BLEU & Ru-BLEU & De-BLEU & Fi-BERTScore & Ru-BERTScore & De-BERTScore \\
\hline
\textit{TartuNLP (ours)}    & 0.028 & \textbf{0.587} & \textbf{0.01} & 0.679 & \textbf{0.869} & 0.63 \\
WooperNLP   & 0.023 & 0.027 & \textbf{0.01} & 0.675 & 0.656 & \textbf{0.65} \\
ABDN-NLP    & \textbf{0.107} & 0.027 & 0.0  & \textbf{0.706} & 0.677 & 0.0 \\
baseline    & 0.033 & 0.005 & 0.0  & 0.403 & 0.377 & 0.49 \\
\hline
\end{tabular}
\caption{Language specific results for the Subtask 2.}
\label{tab:lang-specific}
\end{table*}

\section{Results}

The overall results of both subtasks are presented in Tables~\ref{tab:task1-all} and ~\ref{tab:task2-all}.
For subtask 1, the metrics reported are the average macro-F1 score and the average Adjusted Rand Index (ARI)~\cite{hubert1985comparing} across target words per language. The overall F1 and ARI scores are computed as the mean across all languages. 
For subtask 2, the evaluation metrics are the BERTScore~\cite{zhang2020bertscore} and  BLEU~\cite{papineni-etal-2002-bleu} averaged across target words per language. BLEU and BERTScore values for the entire subtask are the respective averages across all languages. The overall score is the mean of BLEU and BERTScore.

Our submission attained the third place out of eight participants in the first subtask and the first place out of four participants in the second subtask (Table~\ref{tab:task2-all}). This aligns with our expectations since we focused on the second subtask from the beginning and applied the system developed for the second subtask to the first subtask. 
When looking at the language-specific measures of subtask 2 (Table~\ref{tab:lang-specific}), one can see considerable differences between languages. Our system works the best in Russian while also performing well in German in terms of BERTScore (although the BLEU score is close to 0 for all systems). In Finnish, our system is competitive in terms of BERTScore but underperforms compared to the baseline in terms of BLEU.

\section{Discussion}

Our submission to the second subtask is well ahead of the other participants in the overall leaderboard (Table~\ref{tab:task2-all}) despite the simplicity of our approach. However, the language-specific results show that it is not so clear-cut (Table \ref{tab:lang-specific}). Some of the success can be attributed to accidentally matching the source of definitions for the Russian language, which is the Russian Wiktionary. We believe so because the value of the BLEU metric of our submission in the Russian language is higher than that of the other teams and in the other languages by an order of magnitude. However, we do not consider this a critical issue because the BERTScore metric is reasonably high and well above the baseline for all languages, suggesting that the matched definitions capture the expected senses well. However, the corresponding low BLEU scores highlight the inadequacy of the BLEU metric for this task.

Secondly, our approach to the second subtask has limitations. More specifically, matching usage examples only against the definitions of the target word, while efficient, considerably limits the system's ability to describe completely new senses. Intuitively, a definition associated with a different word may be a more suitable description of a new sense. A more robust solution would involve matching usage examples against all available definitions. However, that would likely require using a bi-encoder architecture (as proposed by~\citet{blevins-zettlemoyer-2020-moving}, for instance) instead of a cross-encoder due to the computational complexity of matching every example with every definition.

\begin{table}[t]
\small
\centering
\begin{tabular}{lc}
\toprule
\textbf{Wiktionary language} & \textbf{Number of unique pages} \\ 
\midrule
Finnish   & 586,439 \\
German   & 1,314,597 \\
Russian   & 2,877,010  \\
\bottomrule
\end{tabular}
\caption{The number of unique Wiktionary pages per language.}
\label{tab:wiki-pages}
\end{table}

Accessing the definitions for all the words in a given language-specific Wiktionary is time-consuming because the layout, article structure, and templates used are completely different for each Wiktionary version. While there is a resource providing Wiktionary dumps in a much more convenient format,\footnote{\url{https://kaikki.org/}} it is mostly limited in its support to the English language, with the support for some languages, such as Russian and German, being work in progress, and for others, such as Finnish, completely missing at the time of writing. Moreover, the Finnish, German, and Russian Wiktionaries differ in size and the fullness of their coverage. A rough estimate can be made by accessing the \textbf{Special:Statistics} page of each Wiktionary and examining the total number of unique pages (Table \ref{tab:wiki-pages}). We note the correlation between the smaller sizes of the Finnish and German Wiktionaries and the lower performance of our system on these languages.

Lastly, although we did not focus on the first subtask, we believe the results of the sense prediction task obtained with our systems could also be improved. For instance,  the choice of the threshold value for determining a new sense could be done in a more systematic manner or made learnable. Similarly, adjusting the training data or the hyperparameters might bring further improvements.

\section{Conclusion}

This paper described our solution to both subtasks of the AXOLOTL-24 shared task based on leveraging classifier probabilities for usage example/sense definition pairs. 
The developed system is conceptually simple, adopting a binary classification approach to predict the probability of a sense definition matching the usage example and employing the adapters framework to reduce computation resource requirements. 
Our submission attained the third place in the first subtask and the first place in the second subtask, showing the feasibility of our approach.

\section*{Acknowledgements}

This research was supported by the Estonian Research Council Grant PSG721.

\bibliography{custom}




\newpage

\appendix

\onecolumn
\section{Training Examples}
\label{appendix}
Table~\ref{tab:training-ex} presents two training examples for the Russian word \fontencoding{T2A}\selectfont ``Перо'' \fontencoding{T1}\selectfont(\textit{Feather}). In the first row, we have a gloss and a matching usage example for the figurative meaning of the word (\textit{a symbol of the writer’s art}), which is denoted by the label \textbf{1}. Each usage example of the word in its other senses is paired with this gloss and used as a negative example labeled \textbf{0}. For instance, in the second row, the same gloss is paired with a mismatching usage example in the literal sense of the word. We omit the rest of the negative examples and the other senses for brevity.

\begin{table}[h!]
\centering
\small
\centering
\begin{tabular}{lcc}
\toprule
\textbf{Gloss} & \textbf{Usage example} & \textbf{Label} \\ 
\midrule
\fontencoding{T2A}\selectfont ``Символ искусства писателя, писательского труда, его ремесла.''   & \fontencoding{T2A}\selectfont ``У него бойкое, острое перо.''  & 1\\
\fontencoding{T2A}\selectfont ``Символ искусства писателя, писательского труда, его ремесла.''   & \fontencoding{T2A}\selectfont ``Перья зверя.'' & 0 \\
\bottomrule
\end{tabular}
\caption{A subset of training examples for a single word.}
\label{tab:training-ex}
\end{table}

\end{document}